\documentclass{article}

\usepackage{times}
\usepackage{graphicx}
\usepackage{subfigure} 

\usepackage{natbib}

\usepackage{algorithm}
\usepackage{algorithmic}

\usepackage{hyperref}

\usepackage[accepted]{icml2015}

\begin{document} 

\twocolumn[
\icmltitle{Boosting Convolutional Features for Robust Object Proposals}

\icmlauthor{Nikolaos Karianakis}{nikos.karianakis@gmail.com}
\icmladdress{University of California, Los Angeles, USA}
\icmlauthor{Thomas J. Fuchs}{thomas.fuchs@jpl.nasa.gov}
\icmladdress{Jet Propulsion Laboratory, California Institute of Technology, Pasadena, USA}
\icmlauthor{Stefano Soatto}{soatto@cs.ucla.edu}
\icmladdress{University of California, Los Angeles, USA}

\icmlkeywords{generic object detection, deep learning, boosting}

\vskip 0.3in
]

\begin{abstract}
Deep Convolutional Neural Networks (CNNs) have demonstrated excellent performance in image classification, 
but still show room for improvement in object-detection tasks with many categories, in particular for cluttered scenes and 
occlusion. Modern detection algorithms like Regions with CNNs \cite{girshick14} rely on Selective Search 
\cite{uijlings13} to propose regions which with high probability represent \emph{objects}, where in turn 
CNNs are deployed for classification. Selective Search represents a family of sophisticated algorithms 
that are engineered with multiple segmentation, appearance and saliency cues, typically coming with a significant run-time 
overhead. Furthermore, \cite{hosang14} have shown that most methods suffer from 
low reproducibility due to unstable superpixels, even for slight image perturbations.
Although CNNs are subsequently used for classification in top-performing object-detection pipelines, current proposal methods 
are agnostic to how these models parse objects and their rich learned representations.
As a result they may propose regions which may not resemble high-level objects or totally miss some of them. 
To overcome these drawbacks we propose a boosting approach 
which directly takes advantage of hierarchical CNN features for detecting regions of interest fast. 
We demonstrate its performance on ImageNet $2013$ detection benchmark and compare it with state-of-the-art methods.
\end{abstract}

\section{Introduction}
\label{intro}

Visual object detection is at the heart of many applications in science and engineering, ranging from microscopic scales 
in medicine to macroscopic scale in space exploration. Historically these problems are mostly formulated as classification 
problems in a sliding-window framework. To this end a classifier is trained to differentiate one or more object classes from a 
background class by sliding a window over the whole image with a given stride and on multiple scales and aspect ratios 
predicting the class label for every single window. In practice, this demands the classification of more than a million 
windows for common images, resulting in a trade-off between run-time performance and classification accuracy 
for real-time applications.

Recent years have seen a paradigm shift in object-detection systems replacing the sliding window step with object-proposal 
algorithms prior to classification \cite{girshick14, szegedy14scalable, cinbis13, wang13,  he14}. These algorithms propose regions in an image which are 
predicted to contain the objects of interest with high likelihood. Such an approach not only reduces the number of regions which 
have to be classified from a million to a few thousands but also allows for spanning across a larger range of scales and aspect 
ratios for regions of interest. Depending on the execution time of the proposal method and the classifier, the reduction 
of regions for classification can lead to a significantly faster run time and allow the use of more sophisticated classifiers.

Frameworks with region proposal methods and subsequent classification which are based on Convolutional Neural 
Networks \cite{girshick14} achieve state-of-the-art performance on the ImageNet \cite{deng09} detection. 
A popular pipeline consists of three main steps: (i) several regions that are likely to be objects are generated by a proposal 
algorithm; (ii) deep CNNs with multi-way Softmax or Support Vector Machines (SVMs) on the top classify these regions in order to 
detect all possible classes and instances per image; (iii) finally, an optional regression step further refines the location of 
the detected objects.

A large body of work of region proposal algorithms exists \cite{hosang14}. The majority of them are engineered to merge 
different segmentation, saliency and appearance cues, and some hierarchical scheme, with parameters tuned to specific 
datasets. Here we propose a data-driven region-proposal method which is based on features extracted directly from lower 
convolutional layers of a CNN. We use a fast binary boosting framework \cite{appel13} to predict the objectness 
of regions, and finally deploy a regressor which uses features from the upper convolutional layer after pooling to refine the 
localization. The proposed framework achieves the top recall rate on ImageNet $2013$ detection for Intersection over Union 
(IoU) localization between $50-65\%$. Finally, we apply our proposal algorithm instead of Selective Search in the baseline 
Regions-with-CNN pipeline \cite{girshick14} resulting in $8\%$ improvement over the state-of-the-art with a single prediction 
model while achieving considerably faster test time.

In the two following subsections we briefly review region proposal algorithms and present the motivation behind our work. 
In Section \ref{algo} we present the details of our algorithm and in Section \ref{experiments} we show the experimental study and results. In 
Section \ref{imagenet}, we place our algorithm in line with the Regions-with-CNN framework and benchmark its performance on the 
ImageNet $2013$ detection challenge. Finally, in Section \ref{discussion} we state our conclusions and point out directions 
for future research.

\subsection{Prior work}

Most currently leading frameworks use various segmentation methods and engineered cues in hierarchical algorithms 
to merge smaller areas (e.g., superpixels) to larger ones. During this hierarchical process boxes which likely are objects 
are proposed. In turn CNNs are applied as region descriptors, whose receptive fields are rectangular. This relaxes the 
need to perform accurate segmentation. Instead a rectangle around regions that are most probably objects is sufficient.

We briefly review some representative methods which are evaluated in detail in \cite{hosang14}.

\emph{Selective Search} \cite{uijlings13}, which is currently the most popular algorithm, involves no learning. Its features 
and score functions are carefully engineered on Pascal VOC and ILSVRC so that low-level superpixels \cite{felzenszwalb04} are 
gradually merged 
to represent high-level objects in a greedy fashion. It achieves very high localization accuracy due to the initial 
over-segmentation at a time overhead. \emph{RandomizedPrim's} \cite{manen13} is similar to Selective Search in terms 
of features and the process of merging superpixels. However, the weights of the merging function are learned and the 
whole merging process is randomized.

Then there is a family of algorithms which invest significant time in a good high-level segmentation. 
\emph{Constrained Parametric Min-Cuts (CPMC)} \cite{carreira12} generates a set of overlapping segments. Each proposal  
segment is the solution of a binary segmentation problem. Up to $10,000$ segments are generated per image, which are 
subsequently ranked by objectness using a trained regressor. \cite{rantalankila14}, similar in principle to \cite{uijlings13} and 
\cite{carreira12}, merges a large pool of features in a hierarchical way starting from superpixels. It generates several 
segments via seeds like CPMC does.

\cite{endres10, endres14} combine a large set of cues and deploy a hierarchical segmentation scheme. 
Additionally, they learn a regressor to estimate boundaries between surfaces with different orientations. They use graph cuts 
with different seeds and parameters to generate diverse segments similar to CPMC. \emph{ Multiscale Combinatorial Grouping (MCG)} 
\cite{arbelaez14} combines efficient normalized cuts and CPMC \cite{carreira12} and achieves competitive results within a 
reasonable time budget.

In the literature several methods which engineer objectness have been proposed in the past.
\emph{Objectness} \cite{alexe12} was one of the first to be published, although its performance is inferior 
to most modern algorithms \cite{hosang14}. Objectness estimates a score based on a combination 
of multiple cues such as saliency, colour contrast, edge density, location and size statistics, and the overlap of 
proposed regions with superpixels. \cite{rahtu11} improves Objectness by proposing new
 cues and combine them more effectively.

Moreover, fast algorithms with approximate detection have been recently introduced. \emph{Binarized Normed Gradients 
for Objectness (BING)} \cite{cheng14} is a simple linear classifier over edge features and is used in a sliding window manner. 
In stark contrast to most other methods, BING takes on average only 0.2s per image on CPU. \emph{EdgeBoxes} 
\cite{zitnick14} is similar in spirit to BING. A scoring function is evaluated in a sliding window manner, with object boundary 
estimates and features which are obtained via structured decision forests.

Given that engineering an algorithm for specific data is not always a desired strategy, there are recent region-proposal 
algorithms which are data-driven. Data-driven Objectness \cite{kang14} is a practical method, where the likelihood of 
an image corresponding to a scene object is estimated through comparisons with large collections of example object patches. 
This method can prove very effective when the notion of object is not well defined through topology and appearance, 
such as daily activities.

A contemporary work which follows the data-driven path is Scalable, High-quality object detection 
\cite{szegedy14scalable}. After they revisited their Multibox algorithm \cite{erhan14}, they are able to integrate region 
proposals and classification in one step end-to-end. By deploying an ensemble of models with robust loss function and their 
newly introduced \emph{contextual features}, they achieve state-of-the-art performance on the detection task.

\subsection{Motivation}

Recent top-performing approaches on ILSVRC detection are based on hand-crafted methods for region proposals, 
such as Selective Search \cite{uijlings13}. However, although these methods are tuned on this 
benchmark, they miss several objects. For example, Selective Search proposes on average $2,403$ regions per 
image in \cite{girshick14} with $91.6\%$ recall of ground truth objects (for $0.5$ IoU threshold). In that case, even with 
\emph{oracle} classification and subsequent localization, more than $8\%$ of object instances will not be detected. This leaves 
significant room for improvement in future algorithms.

Furthermore, algorithms which build on superpixels can be unstable even for slight image perturbations resulting 
in low generalizations \cite{hosang14} on different data.

Object proposal algorithms which are based on low-level cues are agnostic on how a learned network perceives the class 
of objects in the space of natural images. A CNN which is trained in a supervised manner to recognize $1000$ object 
categories on ILSVRC, has learned a rich set of representations to identify local parts and hierarchically merge them toward certain 
class instances at the top layers. As opposed to Segmentation-based methods which merge segments based on simple binary 
criteria like existence of boundaries and color or not, CNN features ideally span the manifold of natural images which 
is a very small subspace inside a high dimensional feature space.

In practice, Selective Search is not scale-invariant. Nevertheless, it is engineered to work well on ILSVRC and Pascal VOC data 
with careful parameter tuning. To this end, Regions with CNN \cite{girshick14} resizes all images to a width of $500$ pixels to 
serve its purpose. However, a data-driven algorithm which is not constrained to build on superpixels bypasses this step.

Regions with CNN uses a linear regressor after classifying the proposed regions to better localize the bounding boxes 
around the object. For this purpose they deploy \emph{pool-5} CNN features. This techinque can be applied 
to proposals in the first place. Of course, class-specific regressors are applicable only after classification, but neverthess 
generic object regressors can also enhance region localization.

All in all, the motivation of this work is to address the conceptual discontinuity between the object proposal method and the
subsequent classification. Using hand-crafted scores in the proposal stage and applying a convolutional neural network for 
classification results not only 
typically in slow run-time but also in the aforementioned instabilities. The key idea is to utilize the convolutional responses of a 
network whose weights are learned to recognize different object classes also for proposals. Finally, we formulate the problem 
with a boosting framework to guarantee fast execution time.

\begin{figure*}[t]
\centering
\includegraphics[width=0.9\linewidth]{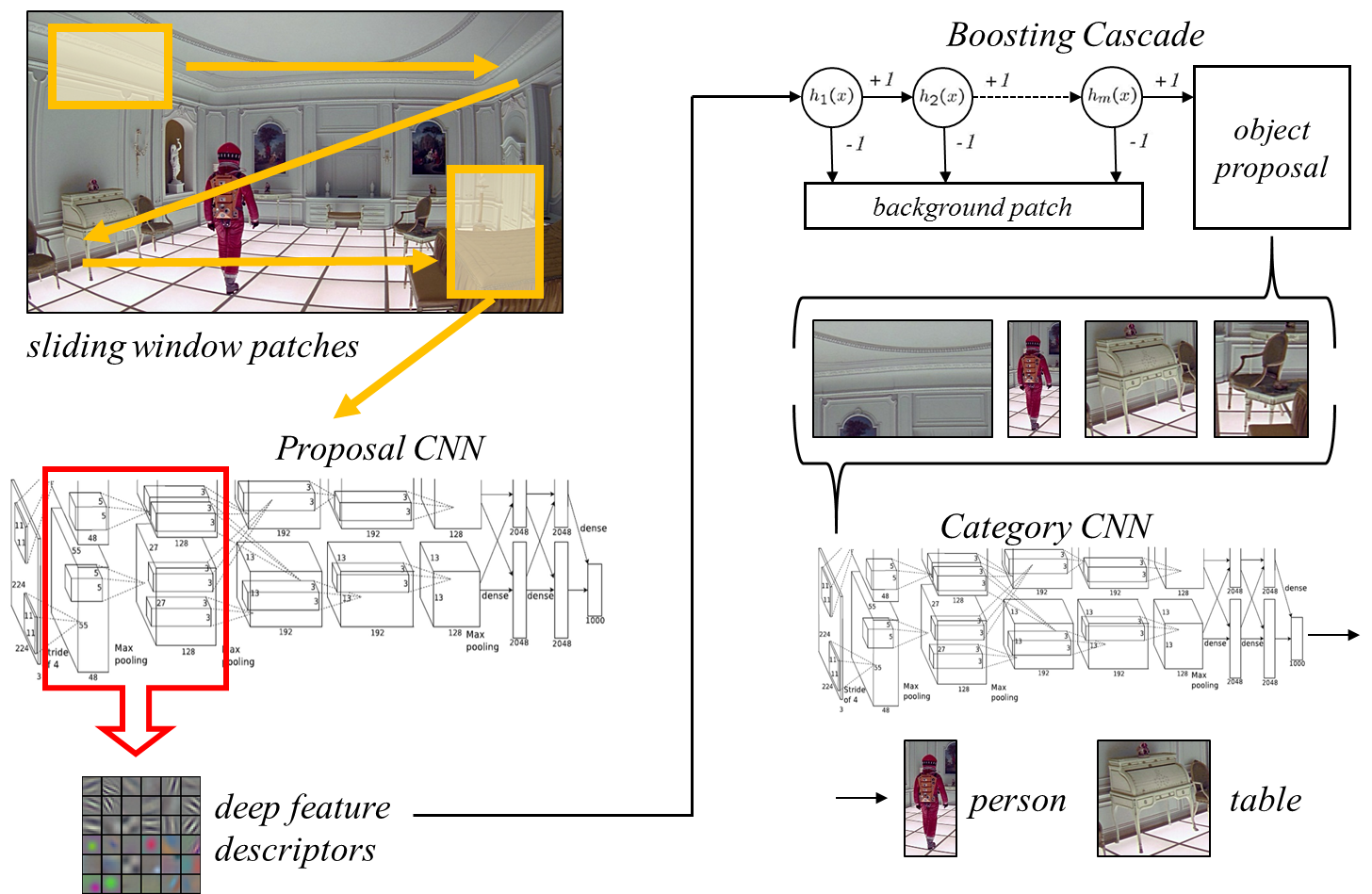}
\caption{Proposed object detection framework based on boosting hierarchical features. Regions which correspond to objects 
and background patches are extracted from training images. Then the convolutional responses from first layers of a Proposal 
CNN are used to describe these patches. These are the input to a boosting model which differentiates object proposals from 
background. Finally a Category CNN is employed to classify the proposals from the Boosting classifier into object categories.}

\label{figure1}
\end{figure*}

\vspace{0.3cm}

\section{Algorithm}
\label{algo}

{\bf Detector:} We are deploying a binary boosting framework to learn a classifier with desired output $y_i \in \{ -1, 1 \}$
for an image patch $i, i \in \{ 1, \ldots, N \}$, where $1$ stands for \emph{object} and $-1$ for \emph{background}. The 
input samples $x_i$ are feature vectors which describe an image patch $i$. The features $x_i$ are a selected subset of 
convolutional responses $conv_j^{k_j}$ from a Proposal CNN (cf. Fig. \ref{figure1}), where $j$ pertains to convolutional layer 
$j \in \{ 1, \ldots, L \}$ and $k_j$ spans the number of feature maps for this layer (e.g., \emph{alexNet} 
\cite{krizhevsky12} uses $L=5$ and $k_1 \in \{ 1, \ldots, 96 \}$). Our Proposal CNN is the \emph{VGGs} model from 
\cite{chatfield14}, whose first-layer convolutional responses are $110 \times 110$ pixels, and therefore provides 
double resolution compared to alexNet.

Aggregate-channel features from \cite{dollar09} are used, where deep convolutional responses serve the role of channels, 
while we deploy a modified version of the fast setting provided by \cite{appel13}. 
Thus, efficient AdaBoost is used to train and combine $2,048$ depth-two trees over the $d \times d \times F$ candidate 
features (channel pixel lookups), where $d$ is the baseline classifier's size and $F$ is the number of convolutional responses, 
i.e., the patch descriptors (e.g., VGGs architecture has $F=96$ kernels in the first layer). The convolutional responses from 
all positive and negative patches which are extracted from the training set are rescaled to a fixed $d \times d$ size (e.g, 
$d=25$) before they serve as input to the classifier. In practice, classifiers with various $d$ can be trained to capture 
different resolutions of these representations. On testing all classifiers are applied to the raw image and their detections are 
aggregated and non-maximally suppressed jointly.

{\bf Hierarchical features:} In order to evaluate \emph{objectness} in different patches, we train the classifier with 
several positive and negative samples, which are extracted from Pascal 2007 VOC \cite{everingham10}. Positive 
samples are the ones that correspond to the ground truth annotated objects, while negatives are defined as 
rectangular samples randomly extracted from the training set at different scales and aspect ratios, which have less than 
$0.3$ Intersection over Union overlap with the positives. For our experiments we considered patches sampled from the 
validation sets of VOC 2007 and ImageNet 2013 detection datasets, since both of them are exhaustively  
annotated\footnote{This means that all object instances belonging to $C$ classes are fully annotated ($C=20$ for Pascal, and 
$C=200$ for ImageNet data), which prevents us from extracting supposedly negative samples which actually correspond to 
objects.}. Naturally there is a margin for improvement with more sophisticated sampling, given that the VOC and ILSVRC  
categories do not include all possible object classes that can appear in an image.

In order to properly crop the objects from the convolutional responses along the hierarchy, the image level annotations have to 
be mapped to the corresponding regions from the intermediate representations effectively. Therefore, the supports of pooling 
and convolutional kernels determine a band within the rectangular box which has to be cropped out, so that information 
outside the object's area can be safely ignored. In practice after two pooling stages the area that should be cropped without 
including too much background information becomes very narrow. In our experiments we consider filter responses from the first 
two layers. We have found that using only kernels from the first convolutional layer, before any spatial pooling is applied, gives 
the best performance.

\begin{figure}[t]

\begin{minipage}[b]{\linewidth}
 \begin{center}
  \includegraphics[width=0.8\linewidth]{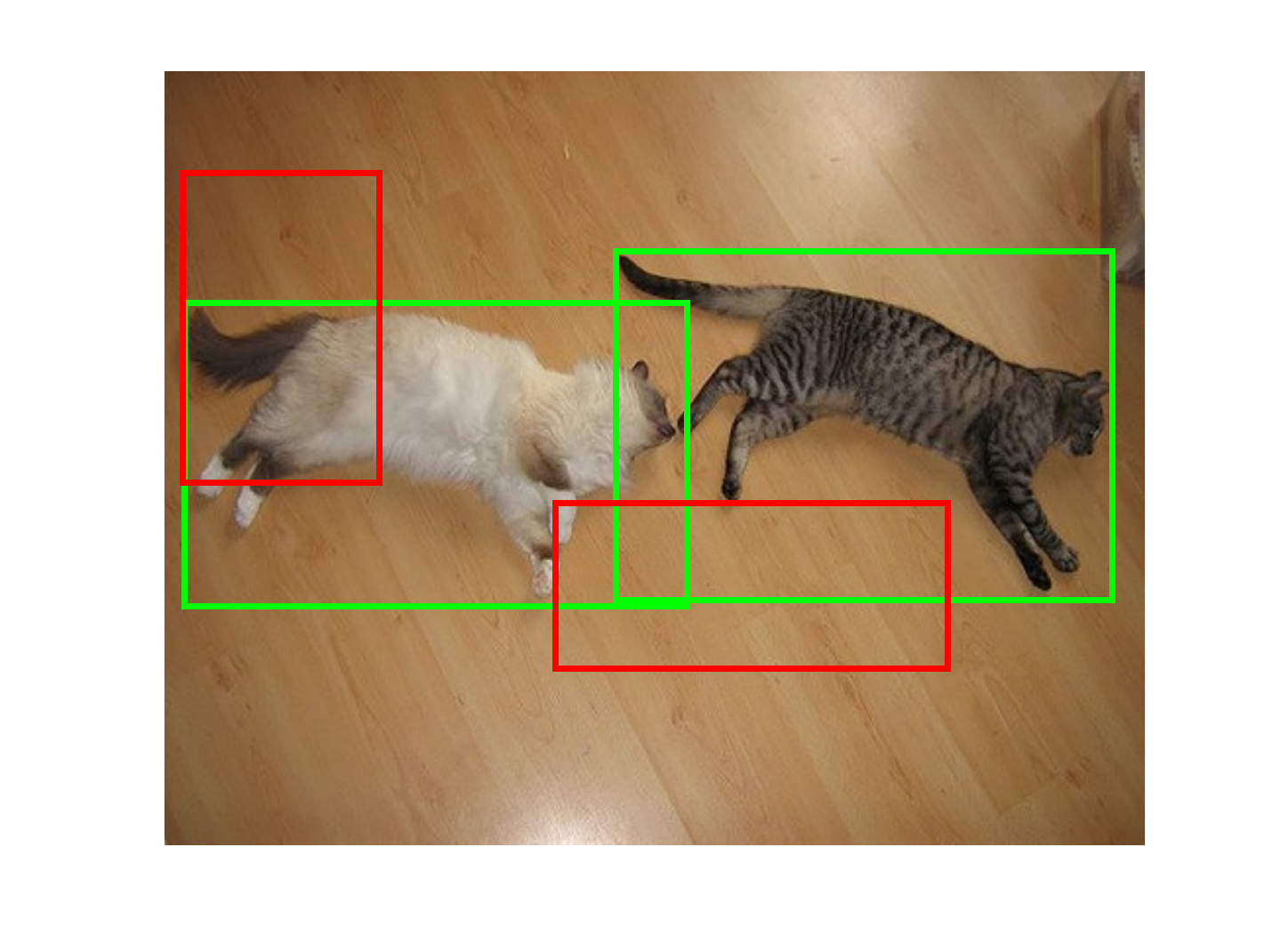}
 \end{center}
\end{minipage}
\begin{minipage}[b]{0.24\linewidth}
  \centerline{\includegraphics[width=0.8cm]{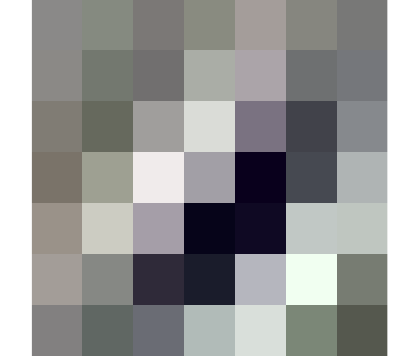}}
\end{minipage}
\hfill
\begin{minipage}[b]{.24\linewidth}
  \centerline{\includegraphics[width=0.8cm]{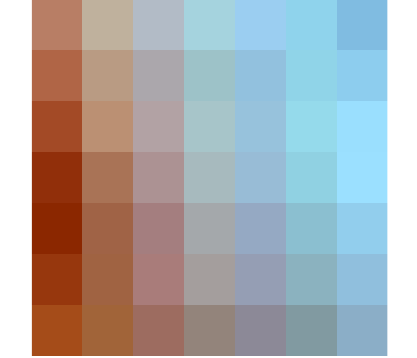}}
\end{minipage}
\hfill
\begin{minipage}[b]{0.24\linewidth}
  \centerline{\includegraphics[width=0.8cm]{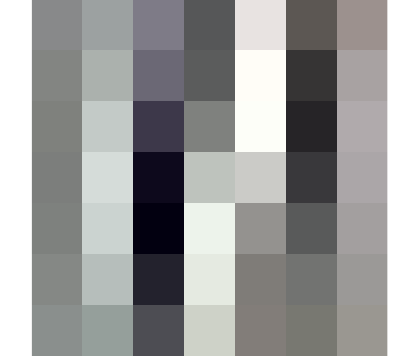}}
\end{minipage}
\hfill
\begin{minipage}[b]{0.24\linewidth}
  \centerline{\includegraphics[width=0.8cm]{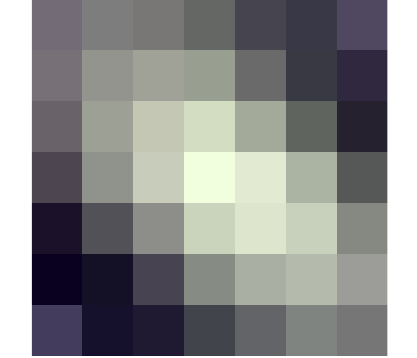}}
\end{minipage}
\vfill
\begin{minipage}[b]{0.24\linewidth}
  \centerline{\includegraphics[width=2.6cm]{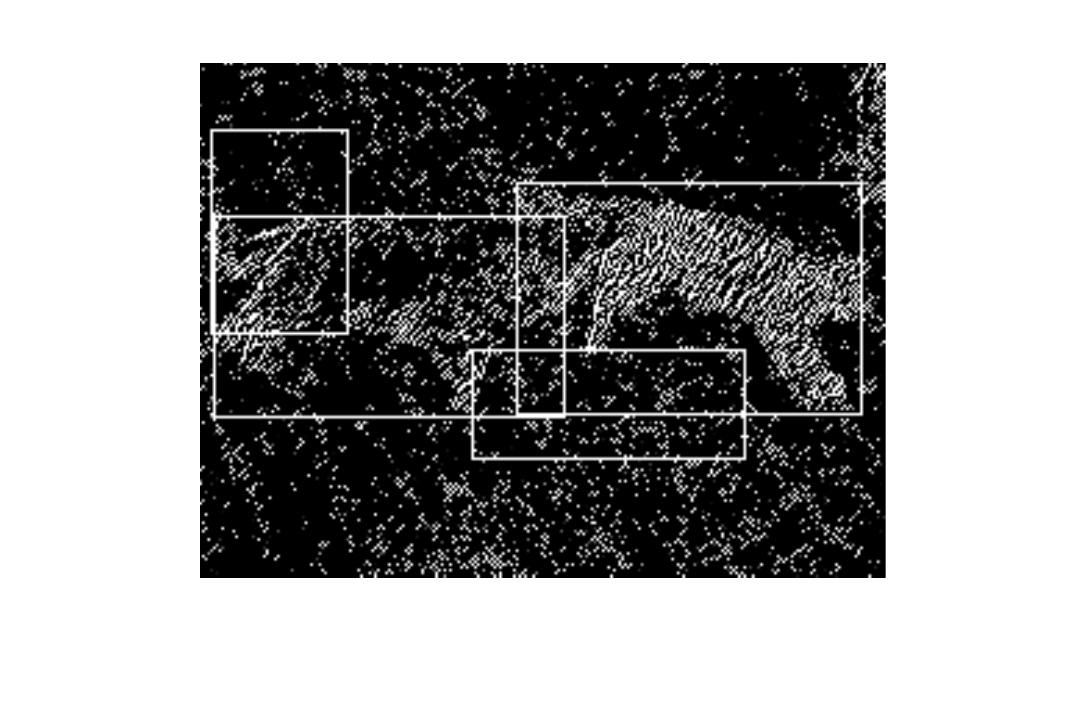}}
\end{minipage}
\hfill
\begin{minipage}[b]{0.24\linewidth}
  \centerline{\includegraphics[width=2.6cm]{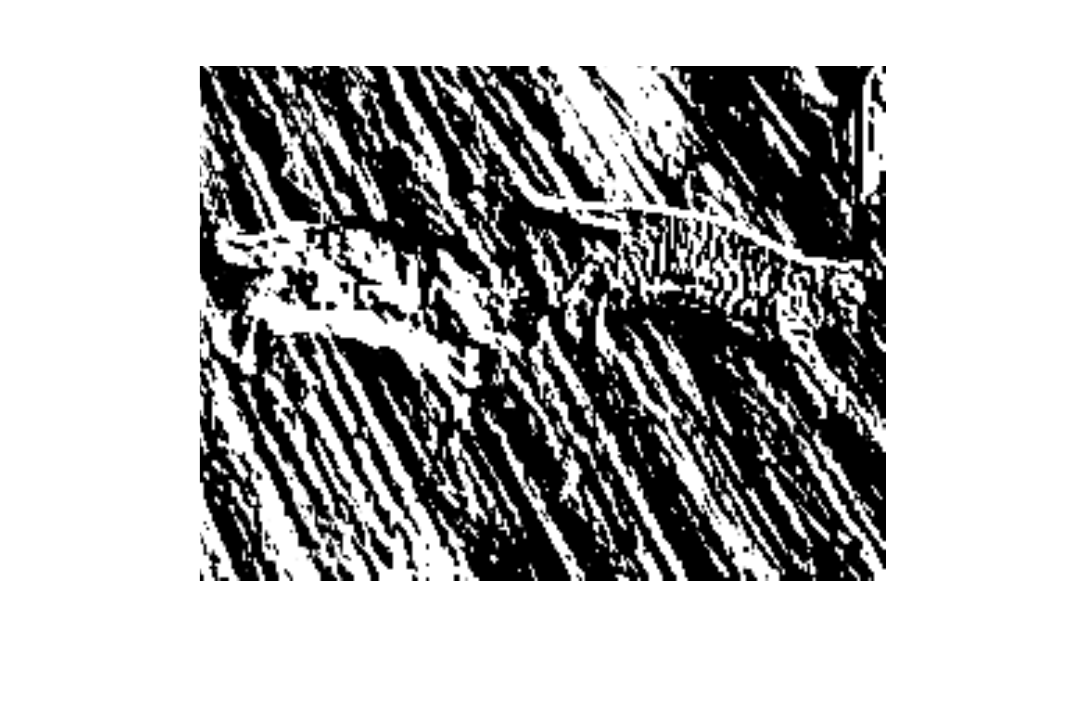}}
\end{minipage}
\hfill
\begin{minipage}[b]{.24\linewidth}
  \centerline{\includegraphics[width=2.6cm]{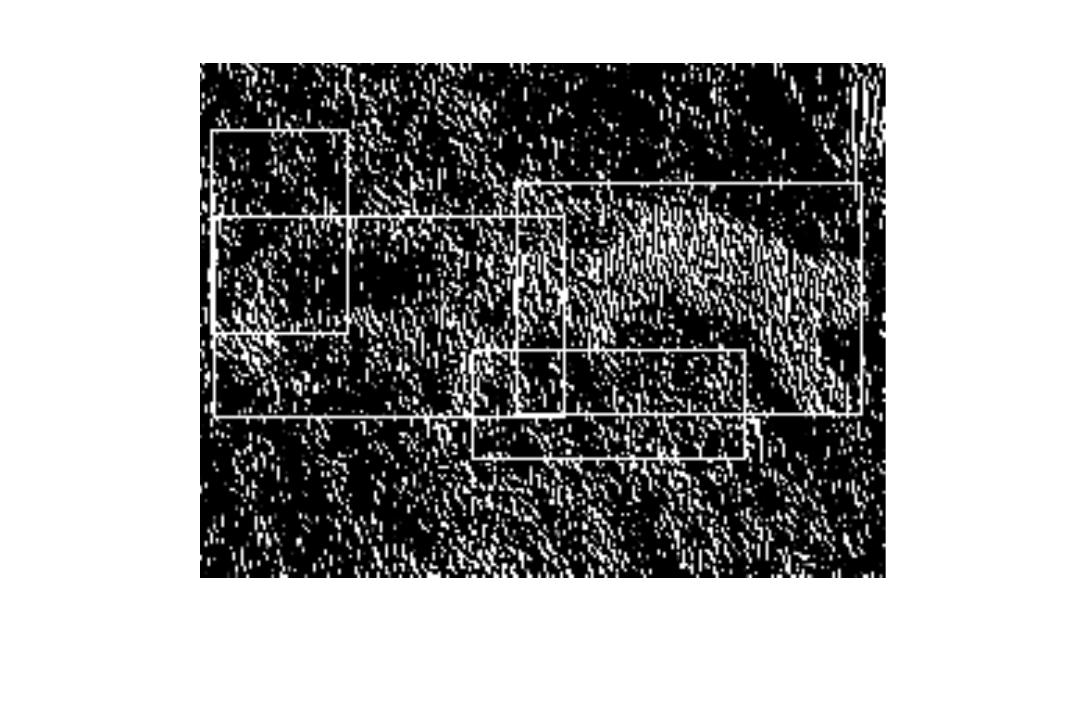}}
\end{minipage}
\hfill
\begin{minipage}[b]{.24\linewidth}
  \centerline{\includegraphics[width=2.6cm]{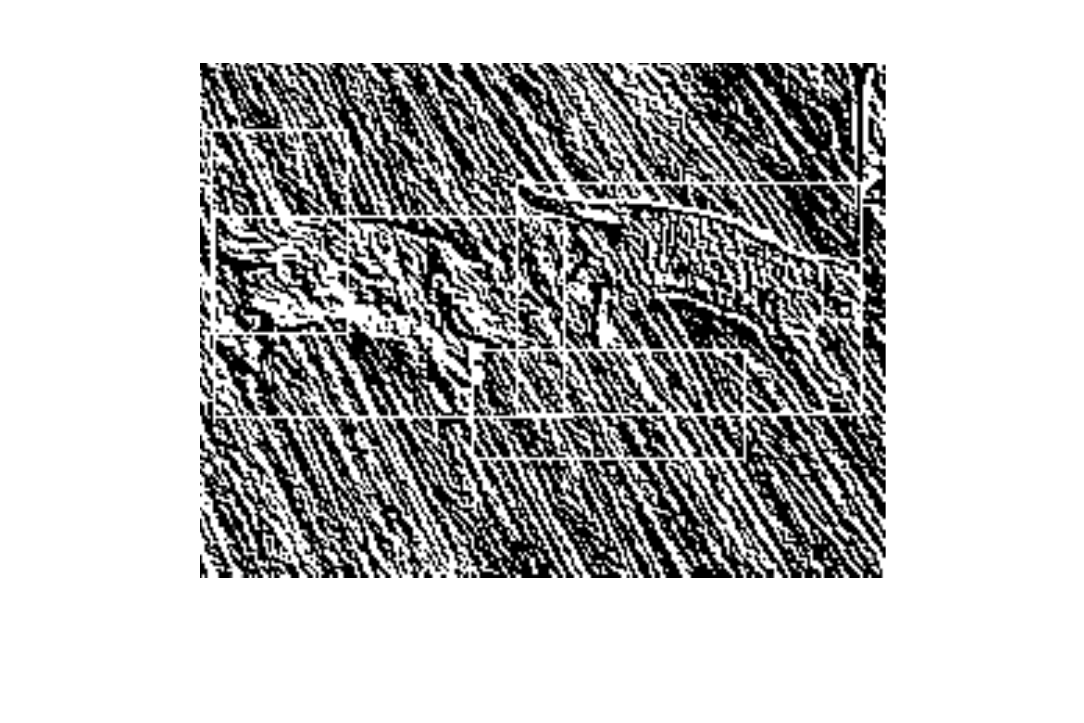}}
\end{minipage}

\caption{An image from Pascal VOC \cite{everingham10} and its convolutional responses with representative first-layer 
filters. In order to classify object candidates a binary boosting framework is trained with positive (green) and negative 
(red) samples which are extracted from CNN's lower layers.}

\label{figure2}
\end{figure}

Most first-layer kernels resemble anisotropic Gaussians and color blobs (cf. Fig. \ref{figure1}). As a matter of fact our method 
relates to BING and EdgeBoxes, which use edge features, and methods that use scores which account for color similarity 
(e.g., Selective Search). By inspecting the convolutional responses of the first layer, we observe that some kernels are able 
to capture the texture of certain objects and thus \emph{disentangle} them from the background and the other objects (such 
as the first and third kernels at Fig. \ref{figure2}). This family of features provides quick detection, as the time-consuming 
high-level segmentation is avoided. However, it has the drawback that local information from lower layers cannot naturally 
compose non-rigid high-level objects. Nevertheless, a subsequent regression step which leverages features from 
the upper convolutional layers can help with these cases. This is described at a following paragraph.

{\bf Testing:} In order to detect objects in previously unseen images we apply the learned classifier in a sliding window 
manner densely in $S$ different scales and $R$ aspect ratios. We typically sample $S=12$ scales and $R=3$ aspect ratios.
Non-maximum suppression (NMS) is used to reject detections with more than $U$ Intersection over Union overlap for every 
(scale, aspect ratio) combination. Finally, after detections from all scales and aspect ratios are aggregated, another joint NMS 
with $V$ IoU is applied. We experimented with different parameters and we use $U=63\%$ and $V=90\%$ in our reported 
results in Fig. \ref{figure3}.

{\bf Bounding-Box Regression:} After extracting the proposals per image, a subsequent regression step can be deployed 
to refine their localization. As proposed by \cite{girshick14}, a linear regressor is used with regularization constant 
$\lambda = 1,000$. For training we use all ground truth annotations $G^i$ and our best detection $P^i$ per ground truth 
for all training images $i, i \in \{ 1, \ldots,  N \}$ from Pascal VOC 2007. The best detection is defined as the one with the 
highest overlap with the ground truth. We throw away pairs with less than $70\%$ IoU overlap. The goal of the regressor is to 
learn how to shift the locations of $P$ towards $G$ given the description of detected bounding box $\phi$.  The transformations 
are modeled as linear functions of $pool_5$ features, which are obtained by forward propagating the $P$ regions through the 
Proposal CNN.

\vspace{0.3cm}

\section{Experiments}
\label{experiments}

\begin{figure*}

\centerline{\includegraphics[width=20cm]{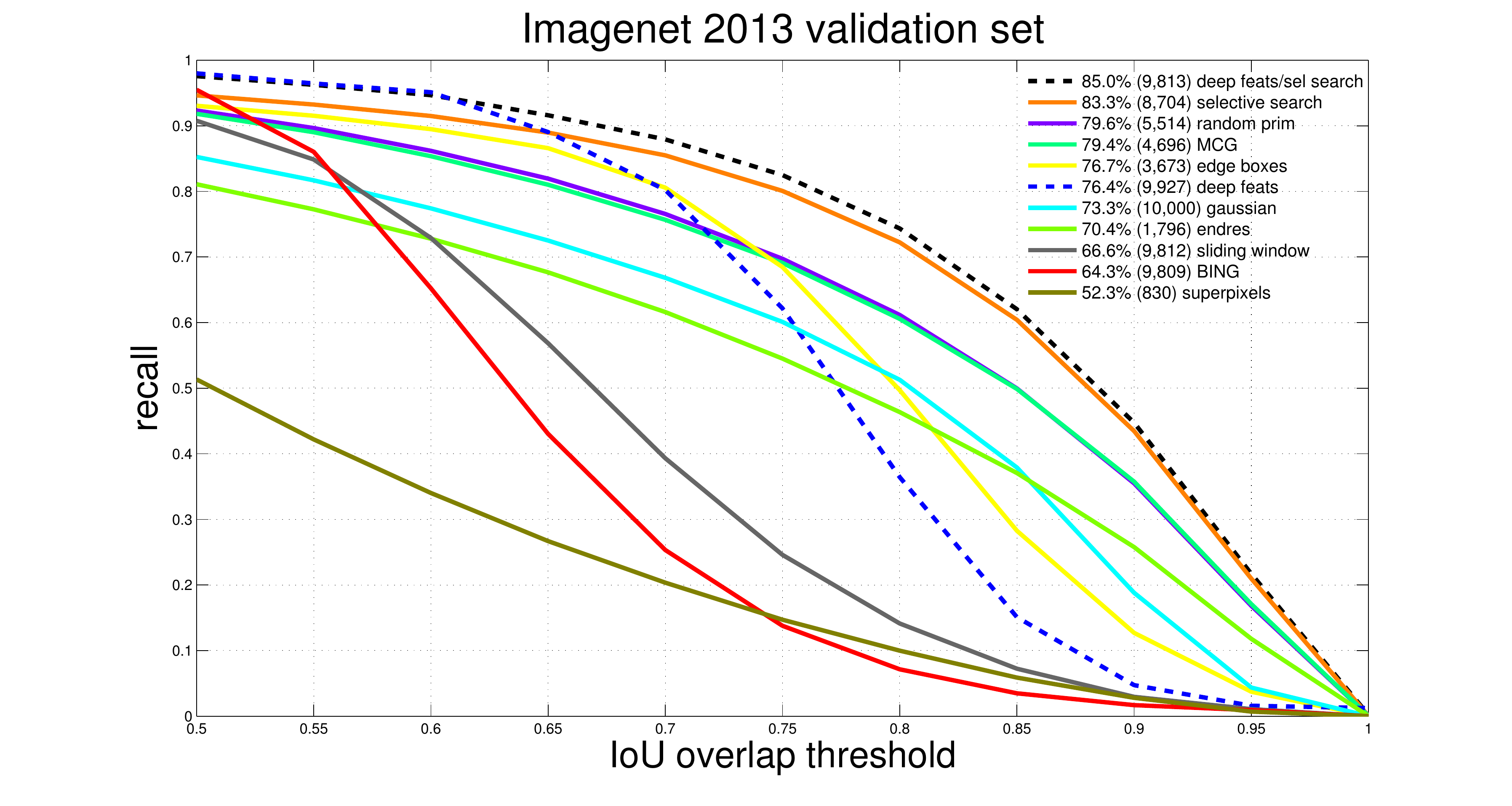}}
\caption{Proposals quality on ImageNet 2013 validation set when at most 10,000 regions are proposed per image. On recall 
versus IoU threshold curves, the number indicates area under the curve (AUC), and the number in parenthesis is the obtained 
average number of proposals per image. Statistics of comparison methods come from \cite{hosang14}. Our curves are drawn dashed.}

\label{figure3}
\end{figure*}

\begin{figure}

\centerline{\includegraphics[width=9cm]{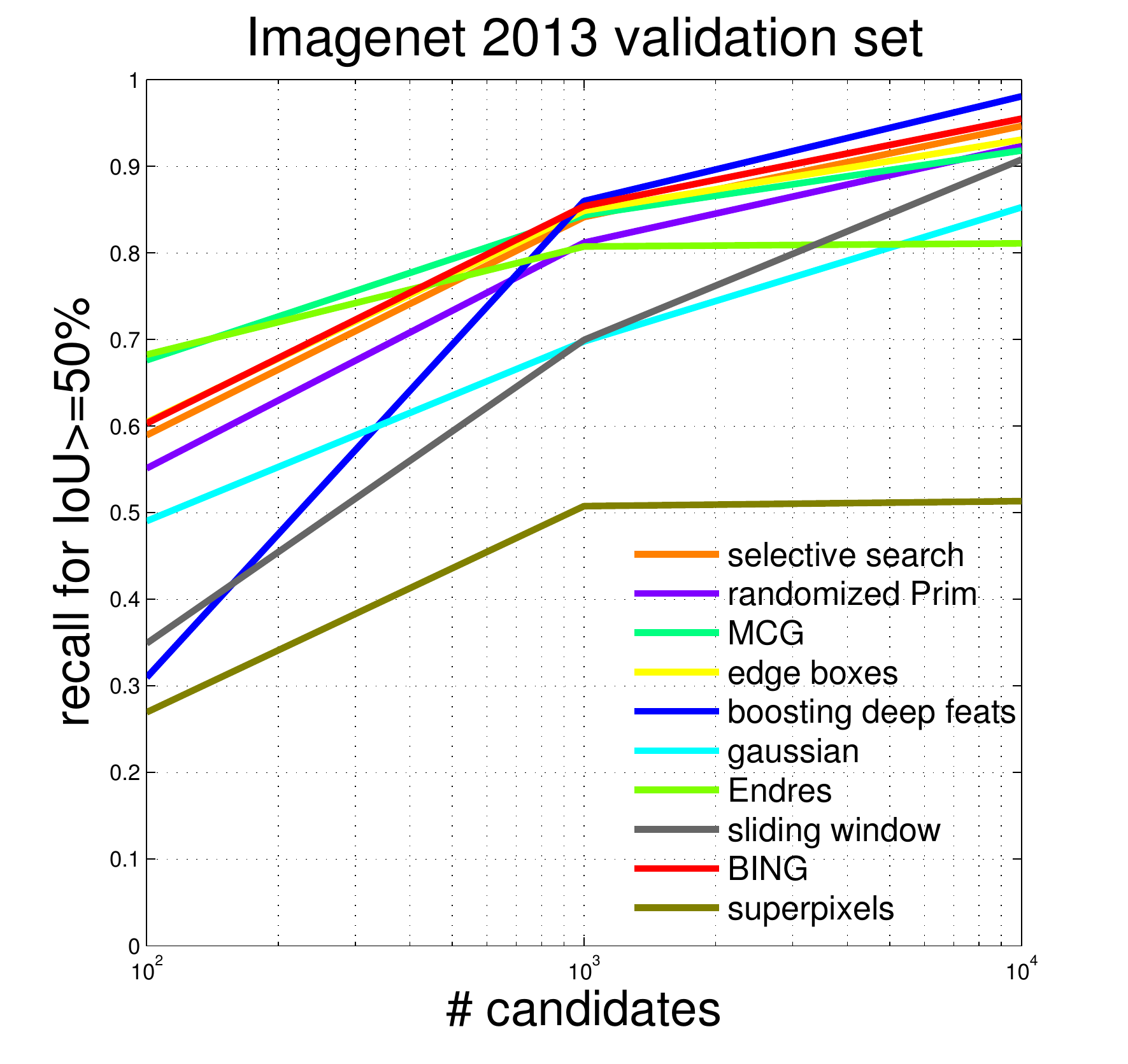}}
\caption{Proposals quality on ImageNet 2013 validation set in terms of detected objects with at least 50\% IoU for 
various average number of candidates per image. Compared to all other methods from \cite{hosang14}, our method 
is the most effective in terms of ground truth object retrieval when at least $1,000$ regions are proposed and 
accurate localization is not a major concern.}

\label{figure4}
\end{figure}

In order to test the efficacy and performance of our algorithm we performed experiments on data from the ImageNet $2013$ 
detection challenge\footnote{As opposed to the training set for both the boosting framework and the regressor, which is Pascal 
$2007$ VOC.} \cite{deng09}. We follow the approach proposed in the review paper of \cite{hosang14} and we report the 
obtained performance vs. localization accuracy (Fig. \ref{figure3}) and number of candidates per image (Fig. \ref{figure4}). 
Specifically, we calculate the recall of ground truth objects for various localization thresholds using the 
Intersection-over-Union (IoU) criterion, which is the standard metric on Pascal VOC. In Fig. \ref{figure3} we demonsrate our 
performance compared to state-of-the-art methods and three baselines, as they have been evaluated by \cite{hosang14}. 
Each algorithm is allowed to propose up to $10,000$ regions per image on average. The methods are sorted based on the 
Area-Under-the-Curve (AUC) metric, while in parentheses is the average number of proposed regions per image.

\begin{table*}

\begin{center}
\begin{tabular}{| c || c | c | c || c | }
\hline Recall ($\%$) for various IoU thresholds & $IoU \geq 0.5$ & $IoU \geq 0.65$ & $IoU \geq 0.8$ & Testing time (s) \\ \hline \hline
Selective Search \cite{uijlings13} & 94.6 & 89.0 & \bf 72.2 & 10\\ \hline
Randomized Prim \cite{manen13} & 92.3 & 82.0 & 61.2 & 1\\ \hline
MCG \cite{arbelaez14} & 91.8 & 81.0 & 60.6 & 30\\ \hline
Edge Boxes \cite{zitnick14} & 93.1 & 86.6 & 49.7 & 0.3\\ \hline
Boosting Convolutional Features & \bf 98.1 & \bf 89.4 & 38.7 & 2\\ \hline
Endres $2010$ \cite{endres10} & 81.1 & 67.7 & 46.4 & 100\\ \hline
BING \cite{cheng14}  & 95.5 & 43.0 & 7.2 & 0.2\\ \hline \hline
Boosting Conv Features and Selective Search & \bf 97.7 & \bf 91.9 & \bf 75.3 & 12\\ \hline \hline
Gaussian & 85.3 & 72.5 & 51.3 & 0\\ \hline
Sliding window & 90.8 & 56.9 & 14.1 & 0\\ \hline
Superpixels & 51.3 & 26.7 & 10.0 & 1\\ \hline
\end{tabular}
\end{center}

\caption{A comparison of our method with various category-independent object detectors on the Validation set of 
ImageNet$2013$ detection data. We compare in terms of recall of ground truth object annotations for various overlap 
thresholds. In order to be consistent with the literature, the strict Intersection-over-Union (IoU) Pascal VOC criterion is used. 
The methods are sorted according to the AUC, similarly to Fig. \ref{figure3}. In bold font the top-2 methods per IoU threshold. 
Representative testing times are shown at the last column.}

\label{table1}
\end{table*}

{\bf Recall vs. localization:} Our method belongs to the family of algorithms with fast and approximate object detection, 
such as BING and EdgeBoxes. These algorithms provide higher recall rate but poorer localization as opposed to methods 
that use  high-level segmentation cues like Selective Search. The latter ones are considerably slower but more 
accurate in localizing the objects due to boundary information. In Table \ref{table1} we provide the recall rate for varying 
localization accuracy via IoU criterion. Our method provides the highest recall until around $65\%$ IoU overlap. We also provide 
in Fig. \ref{figure3} and Table \ref{table1} the gain in performance when we jointly use Selective Search and our method 
while still constraining the number of proposals to be less than $10,000$. In that case the benefit is mutual, as Selective 
Search provides better localization, while our algorithm higher recall, i.e, higher retrieval rate of ground truth objects for 
localization accuracy less than $65\%$ IoU. A small subset of images have been blacklisted in the evaluation process per 
ILSVRC policy.

{\bf Time analysis:} The complexity on testing is linear to each of the number of deployed classifiers, scales, aspect ratios, and 
image size, when the other parameters are held constant. In Table \ref{table1} an estimation of average time on testing 
is shown at the last column for our framework and others, as the latter ones have been evaluated by \cite{hosang14}. 
Our algorithm is quite efficient in both training and testing. More specifically, extracting 
convolutional responses for the validation image set of ImageNet $2013$ detection benchmark takes only a few minutes with 
Caffe \cite{jia14} on a modern machine (e.g., testing one image with a single $K40$ GPU takes $2ms$) mainly because of time 
needed to save the features. However, this can be done offline for popular datasets. Training a modified version of boosting 
framework \cite{appel13} which is now part of Dollar's Matlab toolbox \cite{dollar13} on Pascal VOC 2007 data (train-val and 
test, i.e., 9,963 images with $24,640$ annotated objects) with a high-end multi-core CPU takes about three hours. This 
consists of training on all positives and $20k$ negatives, and additionally three rounds of bootstrapping, when at each round 
$20k$ more negatives are extracted among classifier's false positives. The training time increases for larger values of 
baseline detector's size, such as $d=40$. But this still does not affect the testing time.

When testing the learned classifier is applied densely in sliding window fashion on $20,121$ validation images from ImageNet 
detection. All possible windows in $S=12$ different scales and $R=3$ different aspect ratios are evaluated by the classifier, 
which outputs for each window its confidence to be an object. Greedy non-maximal suppression is performed, where bounding 
boxes are processed in order of decreasing confidence, and once a box is suppressed it can no longer suppress other boxes. 
Separate and joint NMS are deployed with $U=63 \%$ and $V=90 \%$ IoU thresholds, correspondingly. Testing on a 
multi-core CPU takes about $2s$ per image.

In Fig. \ref{figure4} we show performance comparisons in terms of at least $50\%$ IoU recall for different number of proposed 
regions. Our scheme is the most effective when at least $1,000$ regions are proposed. For a smaller number of proposals the
performance degrades fast. This is mainly because of significant non-maximal suppression that is applied to reduce the number 
of proposals, while starting from a large number of scales and aspect ratios. In practice a more sophisticated design for less 
proposals can improve the recall rate further especially in $[100-1,000]$ region. For small number of candidates an alternative 
strategy could be to deploy regression to cluster neighboring regions in unique detection, instead of using non-maximum 
suppression.

\section{ImageNet detection challenge}
\label{imagenet}

In order to investigate in practice how the recall-localization trade-off affects the effectiveness of the object candidates on 
a detection task, we put the challenge to the test and evaluate the overall performance on ImageNet 2013 benchmark. We 
introduce our algorithm into Regions-with-CNN detection framework \cite{girshick14} by replacing Selective Search 
\cite{uijlings13} and proposing our regions instead at the first step.

In Table \ref{table2} we show the mean and median average precision on a subset of validation set. We use the 
$\{ val1, val2 \}$ split as was performed by \cite{girshick14}. We use their pretrained CNN and SVM models as category CNN, 
which are trained on $ \{val_1, train_{1k} \}$, i.e., 9,887 validation images and 1,000 ground truth positives per class from c
lassification set. The deployed Proposal CNN is the \emph{VGGs} model from \cite{chatfield14}, which is pretrained on 
ILSVRC2012 classification dataset. Given that Selective Search is not scale-invariant, all images are rescaled to have $900$ 
pixels width while preserving the aspect ratio. Thus, Selective Search proposes on average $5,826$ regions per image. In 
\cite{girshick14} all images are rescaled to have $500$ pixels width, which yields $29.7$ and $29.2$ mean and median AP for 
$2,403$ regions on average. For our method we used the model that we demonstrate in Figs. \ref{figure3} and \ref{figure4}, 
which generates $9,927$ proposals on average.

\begin{table}[b]

\begin{center}
\begin{tabular}{| c || c | c || c |}
\hline Average Precision (AP) & Mean AP & Median AP\\ \hline \hline
Selective Search & 31.5 & 30.2\\ \hline
Boosting Deep Feats & 34.0 & 32.5\\ \hline
\end{tabular}
\end{center}

\caption{Mean and median average precision on ImageNet 2013 detection task. We deploy the state-of-the-art Regions with 
Convolutional Neural Networks (R-CNN). Comparison when regions are proposed by Selective Search and our 
method, correspondingly. There is no post-processing regression step.}

\label{table2}
\end{table}

Our improvement is a product of two factors; first, higher recall of ground truth objects within roughly $50-70\%$ IoU threshold 
and, second, a larger number of proposals. Coarse localization is corrected to some extent from subsequent steps of R-CNN 
due to the robustness of convolutional neural network in terms of object location and partial occlusion. Further 
improvement is expected if class-specific regression is introduced at the top of prediction, so that the detected bounding boxes 
are better located around the objects, which could prove to be especially helpful in our algorithm given that we leverage no 
boundary/segmentation cues. Finally, an ensemble of models along with more sophisticated architectures (e.g., GoogLeNet 
\cite{szegedy14deep}, MSRA PReLU-nets \cite{he15}, very-deep nets \cite{simonyan14}, etc.) could further improve our results.

\section{Discussion}
\label{discussion}

Features learned in convolutional neural networks have been proven to be very discriminative in recognizing objects in natural 
images by mapping them on small manifolds of a very higher-dimensional feature space. The boosting classifier learns to map 
image data points to the union of all these low-dimensional manifolds. We hypothesize that it is still a relatively small subspace, 
which preserves the notion of object and includes most instances that can be found in popular visual datasets such as Pascal 
VOC and ILSVRC.

In this paper we propose a framework which is able to benefit from the hierarchical and data-adaptive features extracted 
from a convolutional neural network, as well as from the quick training and test time of state-of-the-art boosting algorithms. 
There are many directions to explore this idea further: in this paper the feature responses extracted from different layers 
have equal weight during training, but exploiting the hierarchy in a top-down fashion might result in faster and more accurate 
predictions. Additionally, a data-driven regression mechanism which captures gradient information could improve the localization 
of regions proposed by our framework.

Our framework can also be applied to other areas, such as medical imaging, text detection and planetary science. 
Hand-engineering proposal detectors is quite challenging, as we need to come up with new similarity 
metrics, saliency and segmentation cues. However, instead of designing score functions and features from scratch for each 
new domain, learned deep convolutional features can be used, after a network has been trained 
on a sufficiently large and representative sample of related data. Finally, non-linear tree-based classifiers like boosting 
or random forests can provide a framework for fast inference, while avoiding the overhead of complete propagation in deep 
neural networks and at the same time being flexible to opt for a subset of actionable features for the task.

\nocite{*}

\bibliography{boost_conv_feats}

\begin{thebibliography}{42}
\providecommand{\natexlab}[1]{#1}
\providecommand{\url}[1]{\texttt{#1}}
\expandafter\ifx\csname urlstyle\endcsname\relax
  \providecommand{\doi}[1]{doi: #1}\else
  \providecommand{\doi}{doi: \begingroup \urlstyle{rm}\Url}\fi

\bibitem[Alexe et~al.(2012)Alexe, Deselaers, and Ferrari]{alexe12}
Alexe, B., Deselaers, T., and Ferrari, V.
\newblock Measuring the objectness of image windows.
\newblock \emph{IEEE Transactions on Pattern Analysis and Machine
  Intelligence}, 2012.

\bibitem[Appel et~al.(2013)Appel, Fuchs, Doll\'ar, and Perona]{appel13}
Appel, R., Fuchs, T., Doll\'ar, P., and Perona, P.
\newblock Quickly boosting decision trees-pruning underachieving features
  early.
\newblock In \emph{JMLR}, 2013.

\bibitem[Arbel{\'a}ez et~al.(2014)Arbel{\'a}ez, Pont-Tuset, Barron, Marques,
  and Malik]{arbelaez14}
Arbel{\'a}ez, P., Pont-Tuset, J., Barron, J., Marques, F., and Malik, J.
\newblock Multiscale combinatorial grouping.
\newblock In \emph{Computer Vision and Pattern Recognition}, 2014.

\bibitem[Bergh et~al.(2013)Bergh, Roig, Boix, Manen, and Gool]{bergh13}
Bergh, M. Van~Den, Roig, G., Boix, X., Manen, S., and Gool, L.~Van.
\newblock Online video seeds for temporal window objectness.
\newblock In \emph{International Conference on Computer Vision}, 2013.

\bibitem[Carreira \& Sminchisescu(2012)Carreira and Sminchisescu]{carreira12}
Carreira, J. and Sminchisescu, C.
\newblock {CPMC}: Automatic object segmentation using constrained parametric
  min-cuts.
\newblock \emph{IEEE Transactions on Pattern Analysis and Machine
  Intelligence}, 2012.

\bibitem[Chang et~al.(2011)Chang, Liu, Chen, and Lai]{chang11}
Chang, K.-Y., Liu, T.-L., Chen, H.-T., and Lai, S.-H.
\newblock Fusing generic objectness and visual saliency for salient object
  detection.
\newblock In \emph{International Conference on Computer Vision}, 2011.

\bibitem[Chatfield et~al.(2014)Chatfield, Simonyan, Vedaldi, and
  Zisserman]{chatfield14}
Chatfield, K., Simonyan, K., Vedaldi, A., and Zisserman, A.
\newblock Return of the devil in the details: Delving deep into convolutional
  nets.
\newblock In \emph{British Machine Vision Conference}, 2014.

\bibitem[Cheng et~al.(2014)Cheng, Zhang, Lin, and Torr]{cheng14}
Cheng, M., Zhang, Z., Lin, W., and Torr, P.
\newblock {BING}: Binarized normed gradients for objectness estimation at
  300fps.
\newblock In \emph{Computer Vision and Pattern Recognition}, 2014.

\bibitem[Cinbis et~al.(2013)Cinbis, Verbeek, and Schmid]{cinbis13}
Cinbis, R.~G., Verbeek, J., and Schmid, C.
\newblock Segmentation driven object detection with fisher vectors.
\newblock In \emph{International Conference on Computer Vision}, 2013.

\bibitem[Deng et~al.(2009)Deng, Dong, Socher, Li, Li, and Li]{deng09}
Deng, J., Dong, W., Socher, R., Li, L.~J., Li, K., and Li, Fei-Fei.
\newblock Imagenet: A large-scale hierarchical image database.
\newblock In \emph{Computer Vision and Pattern Recognition}, 2009.

\bibitem[Doll{\'a}r(2013)]{dollar13}
Doll{\'a}r, P.
\newblock Image and video matlab toolbox.
\newblock 2013.

\bibitem[Doll{\'a}r et~al.(2009)Doll{\'a}r, Tu, Perona, and Belongie]{dollar09}
Doll{\'a}r, P., Tu, Z., Perona, P., and Belongie, S.
\newblock Integral channel features.
\newblock In \emph{British Machine Vision Conference}, 2009.

\bibitem[Endres \& Hoiem(2010)Endres and Hoiem]{endres10}
Endres, I. and Hoiem, D.
\newblock Category independent object proposals.
\newblock In \emph{European Conference on Computer Vision}. 2010.

\bibitem[Endres \& Hoiem(2014)Endres and Hoiem]{endres14}
Endres, I. and Hoiem, D.
\newblock Category-independent object proposals with diverse ranking.
\newblock \emph{IEEE Transactions on Pattern Analysis and Machine
  Intelligence}, 2014.

\bibitem[Erhan et~al.(2014)Erhan, Szegedy, Toshev, and Anguelov]{erhan14}
Erhan, D., Szegedy, C., Toshev, A., and Anguelov, D.
\newblock Scalable object detection using deep neural networks.
\newblock In \emph{Computer Vision and Pattern Recognition}, 2014.

\bibitem[Everingham et~al.(2010)Everingham, Gool, Williams, Winn, and
  Zisserman]{everingham10}
Everingham, M., Gool, L.~Van, Williams, C., Winn, J., and Zisserman, A.
\newblock The pascal visual object classes ({VOC}) challenge.
\newblock \emph{International journal of computer vision}, 2010.

\bibitem[Felzenszwalb \& Huttenlocher(2004)Felzenszwalb and
  Huttenlocher]{felzenszwalb04}
Felzenszwalb, P. and Huttenlocher, D.~P.
\newblock Efficient graph-based image segmentation.
\newblock \emph{International Journal of Computer Vision}, 2004.

\bibitem[Feng et~al.(2011)Feng, Wei, Tao, Zhang, and Sun]{feng11}
Feng, J., Wei, Y., Tao, L., Zhang, C., and Sun, J.
\newblock Salient object detection by composition.
\newblock In \emph{International Conference on Computer Vision}, 2011.

\bibitem[Fragkiadaki et~al.(2014)Fragkiadaki, Arbelaez, Felsen, and
  Malik]{fragkiadaki14}
Fragkiadaki, K., Arbelaez, P., Felsen, P., and Malik, J.
\newblock Spatio-temporal moving object proposals.
\newblock \emph{arXiv:1412.6504}, 2014.

\bibitem[Girshick et~al.(2014)Girshick, Donahue, Darrell, and
  Malik]{girshick14}
Girshick, R., Donahue, J., Darrell, T., and Malik, J.
\newblock Rich feature hierarchies for accurate object detection and semantic
  segmentation.
\newblock In \emph{Computer Vision and Pattern Recognition}, 2014.

\bibitem[He et~al.(2014)He, Zhang, Ren, and Sun]{he14}
He, K., Zhang, X., Ren, S., and Sun, J.
\newblock Spatial pyramid pooling in deep convolutional networks for visual
  recognition.
\newblock \emph{European Conference on Computer Vision}, 2014.

\bibitem[He et~al.(2015)He, Zhang, Ren, and Sun]{he15}
He, K., Zhang, X., Ren, S., and Sun, J.
\newblock Delving deep into rectifiers: Surpassing human-level performance on
  imagenet classification.
\newblock \emph{arXiv:1502.01852}, 2015.

\bibitem[Hosang et~al.(2014)Hosang, Benenson, and Schiele]{hosang14}
Hosang, J., Benenson, R., and Schiele, B.
\newblock How good are detection proposals, really?
\newblock In \emph{British Machine Vision Conference}, 2014.

\bibitem[Hosang et~al.(2015)Hosang, Benenson, Doll{\'a}r, and
  Schiele]{hosang15}
Hosang, J., Benenson, R., Doll{\'a}r, P., and Schiele, B.
\newblock What makes for effective detection proposals?
\newblock \emph{arXiv:1502.05082}, 2015.

\bibitem[Humayun et~al.(2014)Humayun, Li, and Rehg]{humayun14}
Humayun, A., Li, F., and Rehg, J.~M.
\newblock {RIGOR}: Reusing inference in graph cuts for generating object
  regions.
\newblock In \emph{Computer Vision and Pattern Recognition}, 2014.

\bibitem[Jia et~al.(2014)Jia, Shelhamer, Donahue, Karayev, Long, Girshick,
  Guadarrama, and Darrell]{jia14}
Jia, Y., Shelhamer, E., Donahue, J., Karayev, S., Long, J., Girshick, R.,
  Guadarrama, S., and Darrell, T.
\newblock Caffe: Convolutional architecture for fast feature embedding.
\newblock In \emph{Proceedings of the ACM International Conference on
  Multimedia}, 2014.

\bibitem[Kang et~al.(2014)Kang, Hebert, Efros, and Kanade]{kang14}
Kang, H., Hebert, M., Efros, A., and Kanade, T.
\newblock Data-driven objectness.
\newblock 2014.

\bibitem[Kr{\"a}henb{\"u}hl \& Koltun(2014)Kr{\"a}henb{\"u}hl and
  Koltun]{krahenbuhl14}
Kr{\"a}henb{\"u}hl, P. and Koltun, V.
\newblock Geodesic object proposals.
\newblock In \emph{European Conference on Computer Vision}. 2014.

\bibitem[Krizhevsky et~al.(2012)Krizhevsky, Sutskever, and
  Hinton]{krizhevsky12}
Krizhevsky, Alex, Sutskever, Ilya, and Hinton, Geoffrey~E.
\newblock Imagenet classification with deep convolutional neural networks.
\newblock In \emph{Advances in neural information processing systems}, 2012.

\bibitem[Manen et~al.(2013)Manen, Guillaumin, and Gool]{manen13}
Manen, S., Guillaumin, M., and Gool, L.~V.
\newblock Prime object proposals with randomized {P}rim's algorithm.
\newblock In \emph{International Conference on Computer Vision}, 2013.

\bibitem[Rahtu et~al.(2011)Rahtu, Kannala, and Blaschko]{rahtu11}
Rahtu, E., Kannala, J., and Blaschko, M.
\newblock Learning a category independent object detection cascade.
\newblock In \emph{International Conference on Computer Vision}, 2011.

\bibitem[Rantalankila et~al.(2014)Rantalankila, Kannala, and
  Rahtu]{rantalankila14}
Rantalankila, P., Kannala, J., and Rahtu, E.
\newblock Generating object segmentation proposals using global and local
  search.
\newblock In \emph{Computer Vision and Pattern Recognition}, 2014.

\bibitem[Simonyan \& Zisserman(2014)Simonyan and Zisserman]{simonyan14}
Simonyan, K. and Zisserman, A.
\newblock Very deep convolutional networks for large-scale image recognition.
\newblock \emph{arXiv:1409.1556}, 2014.

\bibitem[Szegedy et~al.(2014{\natexlab{a}})Szegedy, Liu, Jia, Sermanet, Reed,
  Anguelov, Erhan, Vanhoucke, and Rabinovich]{szegedy14deep}
Szegedy, C., Liu, W., Jia, Y., Sermanet, P., Reed, S., Anguelov, D., Erhan, D.,
  Vanhoucke, V., and Rabinovich, A.
\newblock Going deeper with convolutions.
\newblock \emph{arXiv:1409.4842}, 2014{\natexlab{a}}.

\bibitem[Szegedy et~al.(2014{\natexlab{b}})Szegedy, Reed, Erhan, and
  Anguelov]{szegedy14scalable}
Szegedy, C., Reed, S., Erhan, D., and Anguelov, D.
\newblock Scalable, high-quality object detection.
\newblock \emph{arXiv:1412.1441}, 2014{\natexlab{b}}.

\bibitem[Uijlings et~al.(2013)Uijlings, van~de Sande, Gevers, and
  Smeulders]{uijlings13}
Uijlings, J., van~de Sande, K., Gevers, T., and Smeulders, A.
\newblock Selective search for object recognition.
\newblock \emph{International Journal of Computer Vision}, 2013.

\bibitem[Viola \& Jones(2004)Viola and Jones]{viola04}
Viola, P. and Jones, M.~J.
\newblock Robust real-time face detection.
\newblock \emph{International Journal of Computer Vision}, 2004.

\bibitem[Wang et~al.(2013)Wang, Yang, Zhu, and Lin]{wang13}
Wang, X., Yang, M., Zhu, S., and Lin, Y.
\newblock Regionlets for generic object detection.
\newblock In \emph{International Conference on Computer Vision}, 2013.

\bibitem[Wang et~al.(2014)Wang, Zhang, Lin, Liang, and Zuo]{wang14}
Wang, X., Zhang, L., Lin, L., Liang, Z., and Zuo, W.
\newblock Deep joint task learning for generic object extraction.
\newblock In \emph{Advances in Neural Information Processing Systems}, 2014.

\bibitem[Zehnder et~al.(2008)Zehnder, Koller-Meier, and Gool]{zehnder08}
Zehnder, P., Koller-Meier, E., and Gool, L.~Van.
\newblock An efficient shared multi-class detection cascade.
\newblock In \emph{BMVC}, 2008.

\bibitem[Zhang et~al.(2011)Zhang, Warrell, and Torr]{zhang11}
Zhang, Z., Warrell, J., and Torr, P.
\newblock Proposal generation for object detection using cascaded ranking
  {SVM}s.
\newblock In \emph{Computer Vision and Pattern Recognition}, 2011.

\bibitem[Zitnick \& Doll{\'a}r(2014)Zitnick and Doll{\'a}r]{zitnick14}
Zitnick, C.~L. and Doll{\'a}r, P.
\newblock Edge boxes: Locating object proposals from edges.
\newblock In \emph{European Conference on Computer Vision}. 2014.

\end{thebibliography}
\bibliographystyle{icml2015}

\end{document}